\documentclass{article}
\usepackage{amsmath}
\usepackage{spconf}
\usepackage{hyperref}
\usepackage{enumitem}
\usepackage{dsfont}
\usepackage{comment}

\usepackage{graphicx} 
\usepackage{amssymb} 
\makeatletter 
\newcommand*{\rom}[1]{\expandafter\@slowromancap\romannumeral #1@} 
\setlist{nosep, leftmargin=14pt}
\ninept
\usepackage{mwe} 

\title{FOAA: Flattened Outer Arithmetic Attention for Multimodal Tumor Classification}

\name{Omnia Alwazzan$^{\star \dagger}$,  Ioannis Patras$^{\star \dagger}$, Gregory Slabaugh$^{\star \dagger}$}
\address{$^{\star}$School of Electronic Engineering and Computer Science, Queen Mary University of London, UK\\$^{\dagger}$Queen Mary's Digital Environment Research Institute (DERI), London, UK}

\begin{document}

%
\maketitle
\begin{abstract}
Fusion of multimodal healthcare data holds great promise to provide a holistic view of a patient's health, taking advantage of the complementarity of different modalities while leveraging their correlation.  
This paper proposes a simple and effective approach, inspired by attention, to fuse discriminative features from different modalities. We propose a novel attention mechanism, called Flattened Outer Arithmetic Attention (FOAA), which relies on outer arithmetic operators (addition, subtraction, product, and division) to compute attention scores from keys, queries and values derived from flattened embeddings of each modality. We demonstrate how FOAA can be implemented for self-attention and cross-attention, providing a reusable component in neural network architectures. We evaluate FOAA on two datasets for multimodal tumor classification and achieve state-of-the-art results, and we demonstrate that features enriched by FOAA are superior to those derived from other fusion approaches. The code is at \href{https://github.com/omniaalwazzan/FOAA}{https://github.com/omniaalwazzan/FOAA}

\end{abstract}
\section{Introduction}\label{sec:intro}
With recent progress in multimodal data acquisition in cancer care, there is an increasing opportunity to leverage multiple datatypes to advance diagnosis and prognosis. Heterogeneous data, including imaging and non-imaging data (e.g. gene expression, clinical data), provides different characterization of the same patient and can lead to delivery of personalized treatments. Correspondingly, there has been increased attention in the literature in methods to leverage multiple modalities for downstream tasks. However, the optimal fusion strategy, to take advantage of their complementarity while levering their correlation, remains an open problem. 

Attention-based approaches offer a promising avenue for integrating multimodal data. For example, \cite{lu2021ai} employed shared additive attention across Whole Slide Image (WSI) patches. Attention was used in \cite{schulz2021multimodal} to concatenate features from WSI images and genomic data; however, the results were overshadowed by image-based results. Interestingly, Guan et al. \cite{guan2021predicting} combined three modalities, using self-attention derived from fused features, revealing more correlations between clinical and image features. Similarly, Cui et al. \cite{cui2021co} adjusted the dimensions of clinical graph attention features to match imaging attention features before integration. Yet, such adjustments may compromise performance as in \cite{schulz2021multimodal}. Others \cite{cai2023multimodal}\cite{lu2022m} \cite{chen2021multimodal} use cross-attention to exploit correlations. Despite the significant achievements obtained by attention in current fusion practices, the potential for further improvement to attention scores is often overlooked. We argue that such improvement enables classifiers to fully exploit the complementary interactions between several modalities without compromising performance. 
In this paper, we introduce a novel attention-based fusion method to enhance tumor classification.  
Our work is inspired by \cite{alwazzan2023moab}, who use outer arithmetic operations which are directly applied to embedded features from each modality for healthcare data fusion.  
However, unlike \cite{alwazzan2023moab}, we apply the outer arithmetic operators in an attention mechanism. We demonstrate this attention enriches feature representations and outperforms \cite{alwazzan2023moab} in experimental results. Also, our approach differs from standard attention as it is applied to 
flattened vectors. Our motivation for flattening is two-fold. First, as we are interested in multi-modality fusion, we encode each modality with a separate input head and flatten it into a latent feature vector. Since each modality may have a different dimensionality (e.g. 2D images, 1D clinical data), flattening provides a consistent 1D representation for any modality, simplifying subsequent operations. Second, on this compact flattened representation, it becomes easy to intermingle features, since based on the arithmetic operators, \emph{each} feature in one modality being combined with \emph{all} features of the other modality.  We show this improves extracted feature maps, producing classifiers that better integrate features when operating in multi-modality fusion models. 

We employ our proposed method to solve two different and important medical imaging classification tasks in medical imaging. The first is to classify breast tumors based on MRI images and clinical metadata; the other is to grade brain tumors from histopathology images and gene expression data. In the breast tumor dataset, our experiments show how FOAA is capable of extracting discriminative features from the whole image without patching or segmenting the lesion proving FOAA's effectiveness. In the brain tumor dataset, FOAA succeeded to distinguish similar grades (Grade II, III) from each other providing a new state-of-the-art (SOTA) performance. To this end, we summarise our main contributions as follows:

\begin{itemize}
    \item We propose a Flattened Outer Arithmetic Fusion (FOAA) mechanism which introduces novel ways to compute attention scores with four operations: outer addition (OA), outer product (OP), outer subtraction (OS) and outer division (OD). These operations are simple and reusable.
    \item We employ a self-attention mechanism in FOAA to extract discriminative features from learned feature maps that further improve the performance of classifiers for single modality tasks. 
    \item We apply FOAA in a cross-attention mechanism, achieving state-of-the-art multi-modality fusion on multiple datasets.
\end{itemize}

\begin{figure*}[ht]
\centering
\includegraphics[scale=0.5]{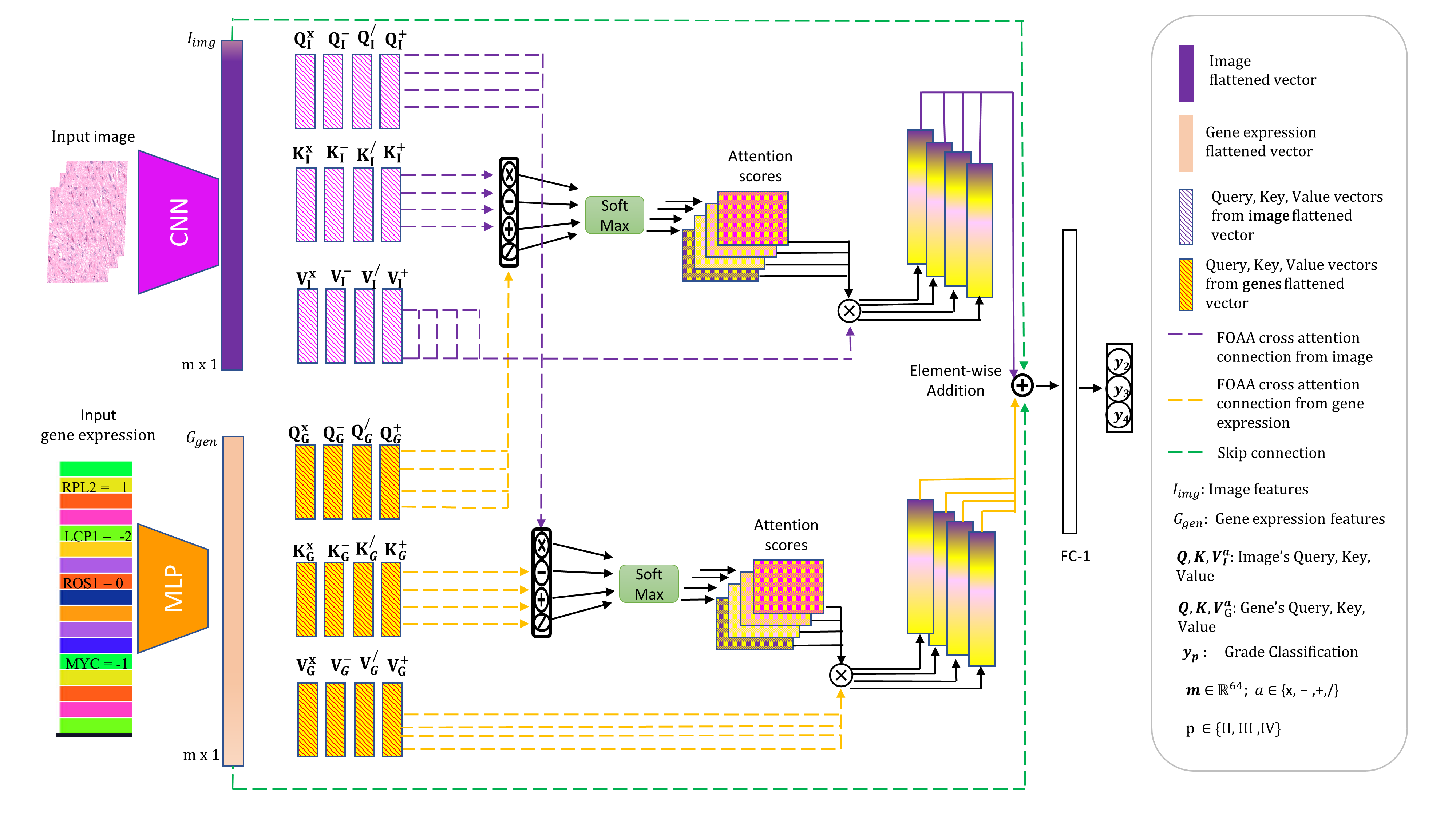}
\caption{The Flattened Outer Arithmetic Attention (FOAA) mechanism is implemented for cross-attention between the image and gene expression modalities occurring in the black box using outer product $\otimes$, outer subtraction $\ominus$, outer division $\oslash$ and outer addition $\oplus$. The resultant attention matrices are applied to the value vector and then integrated with an element-wise sum. This is followed by a fully connected (FC) layer prior to the final classifier.}
\label{fig:fig1}
\end{figure*}

\section{METHOD} \label{sec:method}
The overview of our proposed approach is depicted in Fig.~\ref{fig:fig1} in the context of multi-modality cross-attention fusion. 
The next subsections show how FOAA can be applied in a self-attention mode for unimodal data or in a cross-attention mode for multi-modal data.

\subsection{Unimodal Classifier/Baseline model}
We begin by establishing unimodal baselines by implementing single-modality classifiers. A ConvNeXt-base-v1 \cite{liu2022convnet} convolutional neural network (CNN) was used as a backbone for the imaging modality, and a multilayer layer perceptron (MLP) was used for tabular data. CovnNeXt is a pure convolutional model inspired by the design of Vision Transformers and has been shown to outperform them \cite{liu2022convnet} in various applications. We observed that ConvNeXt is better at extracting features from our medical imaging datasets compared to other backbones such as ResNet18 and VGG19 used in the literature for the same task. Since ConvNeXt has a large depth, we froze all but the last stage (block of layers), as early layers learn the low-level features, whereas the deeper layers learn the high-level features \cite{khan2021image} required for our tasks. By doing so, we were able to train fewer parameters, reduce training time, and accomplish superior performance compared to training from scratch.

\subsection{Flattened Outer Arithmetic Attention (FOAA)}
Inspired by \cite{alwazzan2023moab}, we have adopted the four arithmetic operations; (OA, OP, OS, OD) to replace the scaled-dot product (SDP) in the standard attention mechanism. While \cite{alwazzan2023moab} employs the aforementioned operations to incorporate multiple modalities, we deploy them to compute attention scores. These four operations comprise a block of the attention scores that interrelate features from learned feature maps to further improve the performance of single- and multi-modal classifiers.  
The nature of SDP is to calculate a similarity score between the query $Q$ and each key $K$, which is used to weight the corresponding value $V$. FOAA is different in several ways.  First, the input to our attention is a flattened vector, which allows FOAA to combine every feature in one modality with every feature in the other modality, resulting in a high degree of intermixing of features. Moreover, the unified dimension of extracted feature maps allows versatile and effective integration of distinct modalities.
We acknowledge that the nature of outer products initiates complex computation and acquires more memory in existing methods~\cite{fung2019modality}\cite{hou2019deep}\cite{liu2018efficient}\cite{mallol2022multi}\cite{li2020tpfn}. However, we alleviate this problem using flattened feature maps $m \in \mathbb{R}^{N\times1}$  
where one can control the size of output dimensions to manage computational complexity.

In more detail, we first review the standard scaled dot product equation, formulated as:

\begin{equation} \label{eq1}
\begin{aligned}
s & = f (  \frac{QK^T}{\sqrt{d_k}})V\\
\end{aligned}
\end{equation}
In  Eq.~\ref{eq1} $Q$, $K$, and $V$ are learnable matrices representing queries, keys, and values, respectively, and $d_k$ is the dimensionality of the keys. Here $Q$, $K$, and $V$ $\in \mathbb{R}^{(c,(h\times w))}$ where, $c$ is channel dimension, $h$ and $w$ are height and width size that are flattened to meet attention constraints. The product of the queries and keys, $QK^T$, measures the similarity score $s$ between each query and each key, and scaling the dot product by $\sqrt{d_k}$ helps to stabilise the gradients during training. The softmax function $f$ obtains a probability distribution weighting corresponding values. 

In FOAA, our four ways of computing attention score $s$ can be expressed as the following equations:
\begin{equation} \label{eq2}
\begin{aligned}
OA_{s} & = f\Big( \big( \frac{Q^+\oplus K^{+}}{\sqrt{{d_k}^+}}\bigr)V^+ \Bigr)
\end{aligned}
\end{equation}
\begin{equation} \label{eq3}
\begin{aligned}
OS_{s} & = f\Big( \big( \frac{Q^-\ominus K^{-}}{\sqrt{{d_k}^-}}\bigr)V^- \Bigr)
\end{aligned}
\end{equation}
\begin{equation} \label{eq4}
\begin{aligned}
OP_{s} & = f\Big( \big( \frac{Q^*\otimes K^{*}}{\sqrt{{d_k}^*}}\bigr)V^* \Bigr)
\end{aligned}
\end{equation}
\begin{equation} \label{eq5}
\begin{aligned}
OD_{s} & = f\Big( \big( \frac{Q^\div\oslash K^{\div}}{\sqrt{{d_k}^\div}}\bigr)V^\div \Bigr)
\end{aligned}
\end{equation}
Each of the above equations computes an attention score from the extracted flattened feature vectors producing four enriched vectors that are then combined to make a prediction. The superscript operators $+$, $-$, $*$, and $\div$ represent outer arithmetic operations for each of the weighted scores $OA_{s}$, $OS_{s}$, $OP_{s}$ and $OD_{s}$ $\in\mathbb{R}^{64\times64}$. Note that the queries are derived from the image modality and the clinical modality as part of the cross-attention. For example, given the vectors $Q$ and $K^T$ of the same lengths $m$ $\in \mathbb{R}^{64\times1}$, the outer addition $\oplus$ can be defined as follows:
\[
Q^+ \oplus K^+ =
\begin{bmatrix}
Q_1+K_1 & Q_1+K_2 & \dots  & Q_1+K_m \\
Q_2+K_1 & Q_2+K_2 & \dots  & Q_2+K_m \\
    \vdots & \vdots & \ddots & \vdots \\
Q_m+K_1 & Q_m+K_2 & \dots  & Q_m+K_m
\end{bmatrix}
\]
As illustrated in Fig.~\ref{fig:fig1} we also preserve single-modality features and connect them to the final integrated layer via a skip connection to further boost the model's performance. To determine if FOAA can effectively leverage features specific to a particular modality, we first employed it in a unimodal classifier before a multimodal one. The latter is the key contribution of the paper.

\subsection{Cross-attention FOAA for Multimodal Fusion}
As stated above, our main objective is to apply FOAA to perform cross-attention between multiple modalities. Given two learned flattened feature vectors extracted from their counterpart baseline models, we perform a cross-attention block between these two vectors. As illustrated in Fig.~\ref{fig:fig1} the cross-attention facilitates extensive interaction to combine the fused modalities. This block involves fusing all elements from query $Q$ with all elements from key $K$ that are derived from the image data and clinical data. The four queries $Q$ from each branch are used to exchange information with the key vectors $K$ from the other branch, after which the updated attention scores are projected back to their respective branches.
\subsection{Self-attention FOAA in a Unimodal Classifer}
Similar to our baseline which applies ConvNeXt solely to the image modality, we first extract a vector of 64 features from the input image, then pass it to a self-attention FOAA to obtain its attention scores. However, the difference between our unimodal approach and the pipeline shown in Fig.~\ref{fig:fig1} is that the $K$, $Q$, and $V$ vectors are extracted from the same modality resulting in a self-attention. Then, the enhanced features are aggregated with an element-wise addition operation similar to those in cross-attention module described earlier. 

\section{EXPERIMENTS AND RESULTS}
\label{sec:exp}
\subsection{Dataset} \label{sec:dataset}
We evaluated our proposed approach on two public datasets: A brain tumor dataset derived from the The Cancer Genome Atlas (TCGA), and a breast tumor dataset from the Chinese Mammography Database (CMMD)~\cite{cai2023online}, both providing a supervised learning task.
TCGA is a cancer database containing paired diagnostic whole slide images (WSIs) along with multi-omic data. The brain tumor data was preprocessed and refined by Chen et al.~\cite{chen2020pathomic} providing paired hematoxylin and eosin-stained histopathology images and gene expression data. We note, the majority of WSIs provide slide-level labels that require patching and labelling of the region of interest (ROI); however, Chen et al.~\cite{chen2020pathomic} processing provides patch-level labels reviewed by expert pathologists~\cite{chen2020pathomic}\cite{mobadersany2018predicting}. In astrocytomas and glioblastomas in the merged TCGA-GBM and TCGA-LGG (TCGA-GBMLGG) project, 769 patients represent 1505 histological region-of-interests (ROIs). Each patient's diagnostic WSI provides 1-3 ROIs. Also included are 80 features of the genomic data including 79 copy number variations (CNVs) and one mutation status. The selection and normalization of the genetic features were explored by \cite{mobadersany2018predicting}. Similar to \cite{chen2020pathomic}, we eliminated cases with missing grades, resulting in 396, 408, and 654 ROIs for Grade \rom{2}, \rom{3} and \rom{4}, respectively, that correspond to 736 cases.

The CMMD database provides two views of breast tumors for each patient: craniocaudal (CC) and mediolateral oblique (MLO) views, as well as four other clinical features that can support the imaging modality in our proposed framework. Features are the location of the lesion (as left or right breast), age, lesion subtype (such as mass, calcification, or both), and molecular subtype including luminal A, luminal B, HER2 positive, and Triple-negative. The CMMD comprises 3,728 mammograms from 1,775 cases; 481 of which are benign and 1294 malignant. For 749 of the malignant cases (1,498 mammograms), the molecular subtype of the patients were included; however, we did not use the clinical subtype in our experiments.
\begin{figure}[ht]
\centering
\includegraphics[scale=0.55]{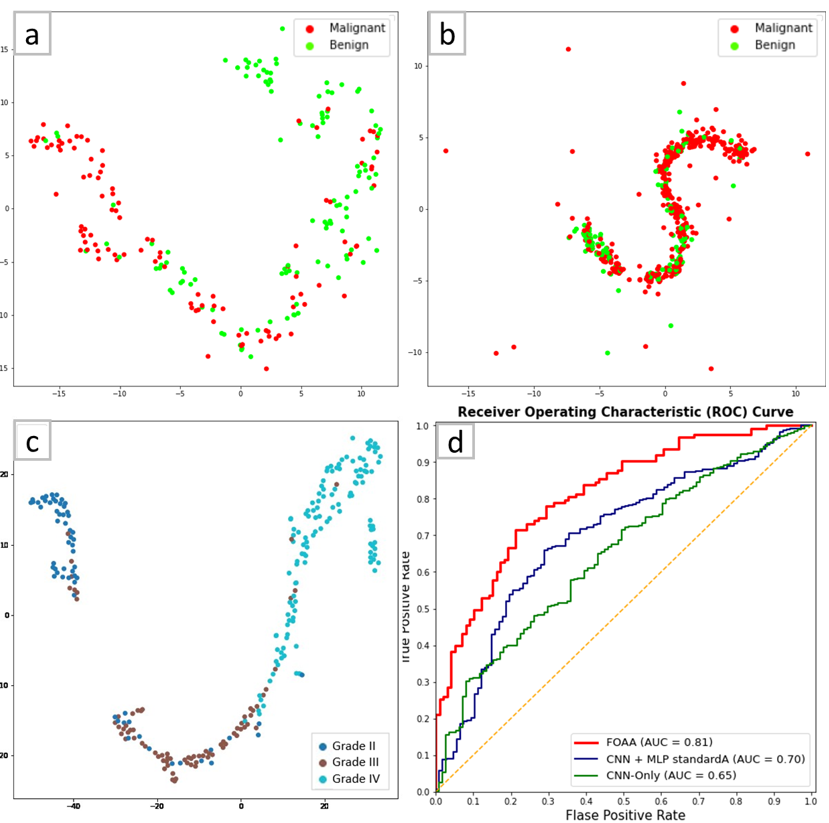}
\caption{t-Distributed Stochastic Neighbor Embedding (t-SNE) visualization for FOAA model. (a) represents FOAA for the CMMD data, (b) replication of MOAB \cite{alwazzan2023moab} on CMMD, (c) FOAA on TCGA, and (d) shows the AUC FOAA in comparison with two ablation studies.}
\label{fig:fig2}
\end{figure}
\subsection{Implementation Details}   \label{sec:Implement}
Similar to \cite{alwazzan2023moab}, our backbone CNN for single and multimodal classification is a ConvNeXt-base-v1 network that is pre-trained on the ImageNet21K dataset. To draw a fair comparison with previous methods conducting tumor grading, we followed an identical image augmentation presented in \cite{chen2020pathomic}\cite{alwazzan2023moab}. Our configuration for the MLP model is also identical to \cite{alwazzan2023moab}. 

For the CMMD data, we first converted the DICOM files into PNG format. Next, we apply the window width and window level stored in the DICOM metadata to adjust the contrast as in\cite{kim2023heterogeneous}. Data augmentation is applied using a histogram equaliser with a probability $=$ 0.4, a random horizontal flip $=$ 0.5 and a random erasing $=$ 0.1. Since the majority of classes are malignant we used a weighted random sampler during training to mitigate the data imbalance. Black pixels outside breast regions are cropped to stabilise training.


Our networks were coded in PyTorch, and implemented using an Adam optimiser, a cross-entropy loss and a batch size of 8. The learning rate was empirically set to 0.00016. To prevent overfitting, our networks were regularised with a weight decay of size 0.005. We chose 64 as the size of the flattened feature vectors involved in FOAA to control computational complexity and memory usage. Each network was trained for 30–40 epochs. A cluster of NVidia A100 GPUs was used for training and inference.

\subsection{Ablation Study} \label{sec:abation}
\begin{table}[ht!]
\caption{Ablation Studies on CMMD}\label{tab1}
    \centering
\begin{tabular}{l|c|c|c|c|c|c}\hline
      Model         & AUC  & Spec & Sens & F1mi & F1ma & Acc \\\hline
      {\footnotesize MLP (metadata)}  &  0.59 & 0.80 &0.30  & 0.50 & 0.50&0.50 \\
      {\footnotesize CNN (image)} & 0.65  & 0.81 & 0.36 & 0.74 & 0.57& 0.74 \\
      {\footnotesize CNN Standard SA} &       0.70 & 0.78 & 0.42 & 0.77  & 0.47 &0.77\\
      {\footnotesize CNN FOAA SA}  &  0.73  & \bf 0.85 &0.42  & 0.72  & 0.64 &0.72\\
      \hline
      {\footnotesize Cross OA} &     0.68  & 0.80 &0.46  & 0.76  & 0.58 &0.76 \\
      {\footnotesize Cross OP} &     0.67  & 0.80 & 0.37 & 0.73  & 0.56 &0.73 \\
      {\footnotesize Cross OS} &      0.72 & 0.82 & 0.45 & 0.78  & 0.59 &0.78	\\		
      {\footnotesize Cross OD} &      0.68 & 0.80 & 0.38 & 0.72  & 0.57 & 0.73 \\			
      {\footnotesize Cross OA+OP}	& 	0.72 & 0.82 & 0.45 & 0.77  & 0.61 & 0.77\\		
      {\footnotesize Cross OA+OP+OS}&0.75 & 0.80 & 0.59 & \bf 0.79  & 0.56 & \bf 0.79 \\
      {\footnotesize MOAB} &          0.70& 0.78 & 0.53 & 0.76 & 0.55&0.76\\
      {\footnotesize \bf FOAA (ours)} &\bf   0.81 &  0.72 & \bf0.76 & 0.74& \bf 0.74&0.74\\\hline
\end{tabular}
\end{table}

\begin{table}[t!]
\caption{TCGA ablation studies and comparison experiments}\label{tab2}
    \centering
    \small 
    \begin{tabular}{l|c|c}
    \hline
         \ Model & F1-Macro &  F1-Micro\\ \hline
        CNN &0.586&0.715\\
        MLP &0.626&0.700\\
        Pathomic \cite{chen2020pathomic}& --   & 0.730  $\pm$ 0.019\\ 
        MultiCoFusion \cite{tan2022multi} & -- & 0.759 $\pm$ 0.032 \\ 
        M2F \cite{lu2022m} & --  & 0.7322  \\ 
        MOAB \cite{alwazzan2023moab} & 0.697  & 0.766   $\pm$0.001\\ \hline
        \bf FOAA (ours) & \bf 0.702&\bf 0.779$\pm$0.024\\ \hline
    \end{tabular}
\end{table}

\begin{table}[ht!]
\caption{CMMD comparison experiments}\label{tab3}
    \centering
    \small 
    \begin{tabular}{l|c|c}
    \hline
         \ Model & AUC &  F1-Macro\\ \hline
        Walsh et al.  \cite{walsh2022comparison}&0.727  $\pm$0.026 & --\\ 
        Miller et al.  \cite{miller2022self} & 0.730 $\pm$0.007  &--  \\ 
        Kim et al. \cite{kim2023heterogeneous} & --  & 0.716 $\pm$0.008  \\
        Gao et al. \cite{gao2023visualize} & 0.771   & --  \\ \hline
        \bf FOAA (ours) & \bf 0.812$\pm$0.002&\bf 0.742$\pm$0.003\\ \hline
    \end{tabular}
\end{table}

We perform a comprehensive ablation study to evaluate FOAA's performance and variants of its design using the CMMD breast tumor data.  We provide numerous evaluation metrics illustrated in Table~\ref{tab1}, namely area under the ROC curve (AUC), specificity (Spec), sensitivity (Sens), F1-score (Micro, Macro), and accuracy (Acc). AUC is our primary result as it provides an aggregate measure of performance across all possible classification thresholds.

We start with baseline unimodal classifiers (top four rows of Table~\ref{tab1}).  Comparing the first two rows, we see that the image data is more discriminative than the metadata. We placed a standard attention (SA) layer after the last conv layer of the ConvNeXt network; then, we combined the updated flattened vector with the MLP flattened vector using element-wise addition. From these results we see FOAA self-attention outperforms standard attention (denoted as CNN FOAA SA and CNN Standard SA in the table).

The remainder of Table~\ref{tab1} presents results of multimodal data fusion. Excluding MOAB, all methods implement cross-attention between the image and metadata. We explore individual outer arithmetic operations, different combinations and FOAA. The model behaves better with the more arithmetic operations we incorporate, proving that FOAA captures effective variations of feature maps, leading to better performance. We can see a steady increment in the AUC when using three variations of attention scores (Cross OA+OP+OS) rather than two (Cross OA+OP). Additionally, we acquire a balanced sensitivity and specificity score when using FOAA which has been hard to obtain in \cite{walsh2022comparison}. Higher sensitivity ensures that cancer cases are not missed, while higher specificity ensures that false positives are minimized, leading to more accurate and meaningful clinical decisions in cancer diagnosis. 

As our arithmetic operation is inspired by \cite{alwazzan2023moab}, we replicated MOAB denoted in Table~\ref{tab1} to fuse the data; still, FOAA outperforms MOAB. Additionally, qualitative results depicted in Fig.~\ref{fig:fig2}(a) show superior class separation compared to MOAB Fig.~\ref{fig:fig2}(b). Note that from Table~\ref{tab1} even when the imaging modality was used in conjunction with the FOAA in a self-attention mode, enhanced features are accumulated, resulting in F1- macro superior to all ablated models.
   
Results on the brain tumor dataset are depicted in Table~\ref{tab3}.  We compare our results to those performing the same classification task, and we utilized identical training protocols. Following a 15-fold Monte Carlo cross-validation training protocol, the reported performance is the mean of the 15 folds. FOAA outperforms existing SOTA methods, as shown in Table~\ref{tab3}. This is also shown in Fig.~\ref{fig:fig2}(c).
In addition, we note that FOAA shows robustness in both imbalanced datasets.

\section{CONCLUSION}
\label{sec:conc}
We introduce a novel fusion framework, called FOAA, for integrating imaging and non-imaging data by employing a novel cross-attention module for the purpose of enhanced classification in healthcare domains. FOAA employs four arithmetic operations to intermingle features, and can be used in both multimodal and unimodal tasks. We validate FOAA using two different multimodal datasets and show that its performance is superior to unimodal classifiers and other SOTA fusion approaches. Due to the generic nature of FOAA, it is a simple and reusable block that can be incorporated into different neural network architectures.  It can also be extended to support a varied range of modalities with minimum edits, which we will explore in future work. 


\section{Acknowledgments}
The authors appreciate the support of the University of Jeddah and the Saudi Arabia Cultural Bureau.  This paper utilised Queen Mary's Andrena HPC facility. This work also acknowledges the support of the National Institute for Health and Care Research Barts Biomedical Research Centre (NIHR203330), a delivery partnership of Barts Health NHS Trust, Queen Mary University of London, St George’s University Hospitals NHS Foundation Trust and St George’s University of London.\label{sec:acknowledgments}
\section{COMPLIANCE WITH ETHICAL STANDARDS}
The paper relies on publicly available data.

\bibliographystyle{IEEEbib.bst}

\bibliography{refs.bib}

\end{document}